\pdfoutput=1

\documentclass[11pt]{article}

\usepackage{acl}

\usepackage{times}
\usepackage{latexsym}

\usepackage[T1]{fontenc}

\usepackage[utf8]{inputenc}

\usepackage{microtype}

\usepackage{inconsolata}

\usepackage{graphicx}

\usepackage{comment}
\usepackage{amsthm}
\usepackage{float}
\usepackage{graphicx}
\usepackage{subfigure}
\usepackage{caption}
\usepackage{amsmath,amssymb}
\usepackage{mathtools}
\usepackage{booktabs}
\usepackage{multirow}
\usepackage{siunitx}
\usepackage{adjustbox}
\usepackage{pifont}
\usepackage{scalerel}
\usepackage{tabularray}
\usepackage{makecell}
\usepackage{wrapfig}

\newcommand{\arr}[1]{\textcolor{black}{#1}}
\newcommand{\lhr}[1]{\textcolor{black}{#1}}
\usepackage{xspace}
\newcommand{\name}{\textit{Privacy Checklist}\xspace}

\definecolor{stepcolor}{HTML}{d79b00}
\definecolor{contentcolor}{HTML}{6c8ebf}

\title{Privacy Checklist: Privacy Violation Detection Grounding on Contextual Integrity Theory}

\author {
    {\bf Haoran Li}\textsuperscript{\rm 1}\thanks{Haoran Li and Wei Fan contributed equally.},
    {\bf Wei Fan}\textsuperscript{\rm 1}\footnotemark[1],
    {\bf Yulin Chen}\textsuperscript{\rm 2},
    {\bf Jiayang Cheng}\textsuperscript{\rm 1}\\
    {\bf Tianshu Chu}\textsuperscript{\rm 3},
    {\bf Xuebing Zhou}\textsuperscript{\rm 3},
    {\bf Peizhao Hu}\textsuperscript{\rm 3},
    {\bf Yangqiu Song}\textsuperscript{\rm 1}\\
    \textsuperscript{\rm 1}HKUST, 
    \textsuperscript{\rm 2}National University of Singapore, 
    \textsuperscript{\rm 3}Huawei Technologies\\
    \texttt{\{hlibt, wfanag, jchengaj\}@connect.ust.hk}, 
    \texttt{chenyulin28@u.nus.edu}\\ 
    \texttt{\{chutianshu3, Xuebing.Zhou, hu.peizhao\}@huawei.com},
    \texttt{yqsong@cse.ust.hk}\\
}

\begin{document}
\maketitle
\begin{abstract}
Privacy research has attracted wide attention as individuals worry that their private data can be easily leaked during interactions with smart devices, social platforms, and AI applications.
Existing works mostly consider privacy attacks and defenses on various sub-fields.
Within each field, various privacy attacks and defenses are studied to address patterns of personally identifiable information (PII).
In this paper, we argue that privacy is not solely about PII patterns.
We ground on the Contextual Integrity (CI) theory which posits that people's perceptions of privacy are highly correlated with the corresponding social context.
Based on such an assumption, we formulate privacy as a reasoning problem rather than naive PII matching. 
We develop the first comprehensive checklist that covers social identities, private attributes, and existing privacy regulations.
Unlike prior works on CI that either cover limited expert annotated norms or model incomplete social context, our proposed privacy checklist uses the whole Health Insurance Portability and Accountability Act of 1996 (HIPAA) as an example, to show that we can resort to large language models (LLMs) to completely cover the HIPAA's regulations.
Additionally, our checklist also gathers expert annotations across multiple ontologies to determine private information including but not limited to PII.
We use our preliminary results on the HIPAA to shed light on future context-centric privacy research to cover more privacy regulations, social norms and standards.
\end{abstract}

\section{Introduction}
\label{sec: intro}

As machine learning models and their applications achieve wide accessibility and applicability, they revolutionize our daily lives.
On the one hand, these data-driven models achieve consistent utility improvements with increasing training corpus.
On the other hand, private data may be collected unintentionally, which leads to individuals' increasing privacy concerns.
To achieve trustworthy AI, great efforts have been made to study privacy issues with regard to emerging applications~\cite{li2023multi, debenedetti2023privacy}.

Existing privacy studies conducted by researchers are relatively fragmented.
Each field has its own privacy formulation based on specific tasks.
For example, in computer networks, network communications' confidentiality is widely discussed~\cite{Nath-2010-middleman, Fonseca-2009-sql} and cryptographic methods are proposed to protect cybersecurity~\cite{Buehrer-2005-sql_inject_prevent, Needham-1978-Using}.
When it comes to deep learning, in both Computer Vision and Natural Language Processing fields, there is a significant body of research on privacy attacks~\cite{Shokri2016MembershipIA, carlini-2021-extracting, li-etal-2023-sentence} and defenses~\cite{Abadi-dpsgd-2016}.

\begin{figure}[t]
\centering
\includegraphics[width=0.998\linewidth]
{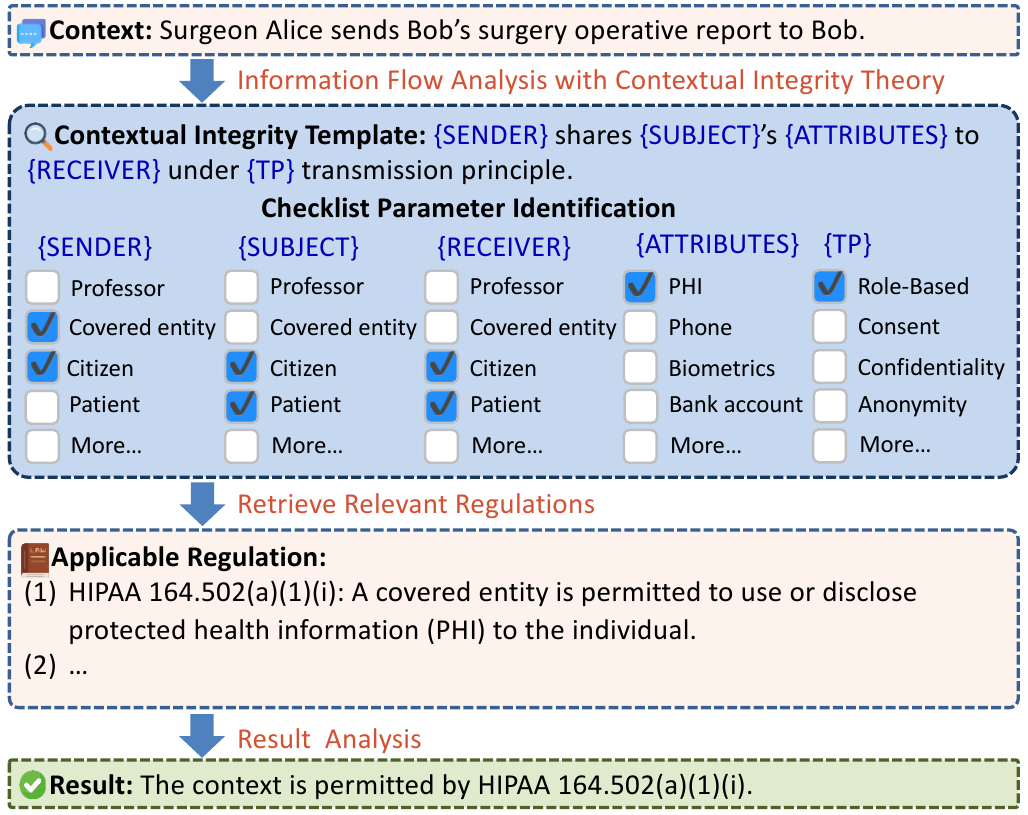}
\vspace{-0.25in}
\caption{
The exemplary case to transform privacy issues into reasoning problems. 
Our proposed checklist collects roles, attributes, transmission principles and annotated legal norms to facilitate the reasoning process.
}
\label{fig:intro}
\vspace{-0.25in}
\end{figure}

In addition to the fragmented formulations, the current scope of privacy research is narrow.
Most existing privacy studies focus on protecting private attributes specified by privacy regulations.
Though these sensitive attributes are indeed crucial components of privacy regulations, they only constitute a small fraction of regulations' coverage.
\lhr{
However,  privacy is not solely about protecting attributes.
Instead, It involves understanding the context to ensure compliance with existing regulations and norms.
}
For example, according to the Health Insurance Portability and Accountability Act of 1996 (HIPAA), protected health information (PHI) is considered sensitive and should be protected.
However, as shown in Figure~\ref{fig:intro}, though Bob's surgery operative report is classified as protected health information, if we identify Alice as a covered entity and Bob as the individual (patient), we may apply HIPAA 164.502(a)(1)(i) to permit surgeon Alice to share Bob's PHI with him.
\lhr{This example demonstrates that privacy issues can be resolved through reasoning within the context. 
When we consider privacy, we actually evaluate whether our information transmission contexts violate existing standards and expectations.
}

To bridge the gap, Contextual Integrity (CI) theory~\cite{Nissenbaum-2010-CI} is proposed to examine the appropriateness of privacy information flows based on specific contexts. 
A few prior works on CI resort to formal languages such as first-order logic~\cite{Barth-2006-CI, Benthall-CI-2017, Shvartzshnaider-2016-Learning} to fit contexts into pre-defined clauses to determine privacy violations.
However, it requires heavy effort for experts to annotate complicated legal documents into logical language.
Consequently, none of these approaches can scale up to cover even a whole document.

In this paper, to leverage the reasoning capability of privacy information flows while allowing substantial flexibility to handle natural language ambiguity and variability, we present \name.
We formulate the privacy evaluation as an in-context reasoning problem with existing large language models (LLMs).
Instead of annotating separate clauses, our \name uses large language models to extract CI characteristics and 
\lhr{use the tree structure to capture their hierarchical nature.
}
In addition, our \name also includes role-based and attribute-based graphs, as well as a definition dictionary to facilitate in-context reasoning.
To determine privacy violations for the given context, we first fetch relevant regulations and use large language models to conduct in-context reasoning with retrieved regulation.
We annotate all information transmission norms inside the HIPAA and show that our \name can improve 6\% - 18\% accuracy on average for existing LLMs to improve their privacy judgment ability in real court cases.
Our contribution can be summarized as follows\footnote{Code is publicly available at \url{https://github.com/HKUST-KnowComp/privacy_checklist}.}:

1)  We extend prior works on CI to natural language and formulate the privacy issue as an in-context reasoning problem with the help of LLMs. 

2) We present \name, a first scalable knowledge base that can cover all norms of information transmission inside the HIPAA.

3) We consider various retrieval-augmented generation (RAG) pipelines. To retrieve relevant legal documents, we implement term frequency, semantic similarity, and agent-based methodologies.

4) We conduct comprehensive experiments to show that our checklist is effective in improving LLMs' privacy judgment for court cases.
\section{Preliminaries}
\label{sec: relate}

\begin{figure*}[t]
\centering
\includegraphics[width=0.999\textwidth]{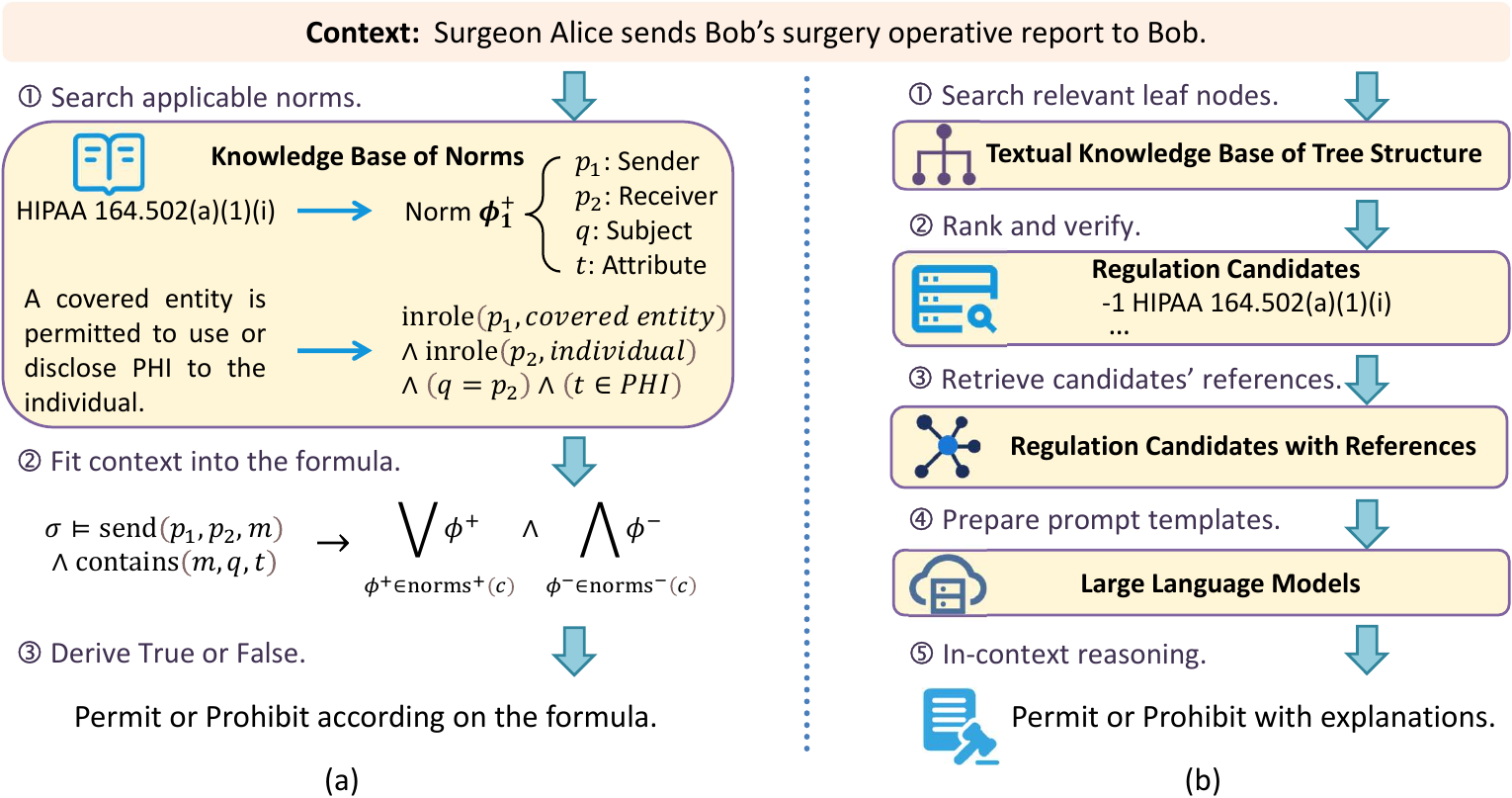}
\vspace{-0.25in}
\caption{
The overview of privacy reasoning within the given contexts. 
Subfigure (a) illustrates previous approaches that use formal languages to determine privacy violations based on rules of inference and axioms. 
Instead, in subfigure (b), we propose an in-context reasoning pipeline with our proposed \name and LLMs.
}
\label{fig:model}
\vspace{-0.2in}
\end{figure*}

\paragraph{Contextual Integrity, norms and applications.}
Contextual Integrity~\cite{Nissenbaum-2010-CI} posits that privacy is related to information flows and information flows must adhere to the established norms of the context to protect privacy. 
Each norm can typically be viewed as an atomic unit of information transmission involving three key actors: the sender, the receiver, and the data subject.
Depending on the social context of these actors, such as their roles and the disclosed attributes of the data subject, we can determine whether the norm is positive or negative, which means the norm is either permitted or prohibited by existing privacy regulations.
Previous works on CI~\cite{Barth-2006-CI, Benthall-CI-2017, Shvartzshnaider-2016-Learning} aimed to transform the context into explicit formal languages, such as first-order logic, to determine privacy violations with rule-based frameworks. 
However, as shown in Figure~\ref{fig:model} (a), it costs heavily for experts to annotate such a knowledge base of norms.
Experts must meticulously define a set of predicates and annotate the entire body of regulations, subpart by subpart.
Consequently, these works on formal languages fail to scale up their norms due to the inherent complexity of legal documents.
\lhr{
On the other hand, existing legal LLMs, including LawGPT~\cite{ zhou2024lawgpt}, Lawyer LLaMA~\cite{huang2023lawyer} and ChatLaw~\cite{cui2023chatlaw} are proposed on the general legal domain and underperform on the legal domain due to data scarcity. 
}
\arr{
GoldCoin \cite{fan2024goldcoin} proposed instruction tuning grounded on CI to enhance LLMs' judgment ability.
Instead, our work reveals the flaws of synthetic data and shows that the proper RAG pipelines without further tuning can yield comparable or even better performance.
}

\arr{
In addition, for discussions about current privacy attacks and defenses, please refer to Appendix~\ref{app: relate}.
}
\section{Privacy Checklist Construction}
\label{sec: checklist}

In this section, we show how to construct our \name based on existing privacy regulations with less annotation cost. 
For the given regulation, our \name encompasses a document tree with CI characteristics, a role graph, an attribute graph, and a definition dictionary.

\subsection{Legal Document Processing}

We start by crawling the full HIPAA document from the official Code of Federal Regulations website.
We implement a parser with regular expressions to capture the regulation identifiers and corresponding contents.
Since the legal documents are rigorously written and well-organized, we can represent the HIPAA rules' subsumption relationships using a tree structure $\mathcal{T} = \{\mathcal{V}, \mathcal{E}\}$ where edges in $\mathcal{E}$ indicates the subsumption relationship between two nodes.
For nodes in $\mathcal{V}$, leaf nodes correspond to specific, non-separable specifications (e.g., 164.502(a)(1)(i)), while internal nodes denote broader subparts (e.g., 164.502(a)(1)). 
We set ``HIPAA'' as the dummy root node.
In addition, we observe frequent cross-references among the nodes.
For each node, any reference of other identifiers is recorded as the node attribute.

\subsection{CI Characteristics Annotation}

For each leaf, we concatenate all texts from the root to the current leaf node to obtain its full specification.
Then, we exploit GPT-4 as the annotator to execute the following tasks:

\paragraph{Specification classification.}
Due to the complexity of legal documents, each leaf node's specification can either be a regulatory norm or a clarification of the norm.
Such clarifications are irrelevant to information transmission but play crucial roles in regulatory norms.
Therefore, for each specification, we ask GPT-4 to conduct a three-way classification to determine if the specification is a positive norm, negative norm, or general definition.
A positive norm denotes the information transmission action (norm) that is permitted by law, while a negative norm prohibits such information transmission.
Otherwise, the specification is regarded as the general definition, which enhances the norms' clarity.

\paragraph{CI characteristics extraction.}
After identifying all the positive and negative norms, we continue to extract characteristics identified by the Contextual Integrity Theory from each norm.
First, we request GPT-4 to identify three actors' roles, including the sender role, receiver role, and subject role for the given norm.
Then, we inquire GPT-4 to determine the subject's attributes that are disclosed during information transmission.
Additionally, we ask GPT-4 to analyze additional context, including the purpose and consent form of transmission.
Finally, we let GPT-4 specify whether the sender and the subject are the same person and, similarly, whether the receiver and the subject are the same person.

\paragraph{Reference annotation.}
Previous works, as shown in Figure~\ref{fig:model} (a), commonly treat each norm as an isolated individual and discard the cross-references among norms.
However, legal regulations are rather complicated, and referring to these references is essential to assessing privacy violations accurately.
For instance, HIPAA 164.502(a)(1)(iv) states that ``A covered entity is permitted to use or disclose protected health information except for uses and disclosures prohibited under 164.502(a)(5)(i), pursuant to and in compliance with a valid authorization under 164.508.''
To permit data transmission with 164.502(a)(1)(iv), we must first ensure that no restrictions apply under 164.502(a)(5)(i) and then confirm that the authorization (one type of consent form) is valid under 164.508.
Therefore, we also extend our checklist to include the annotated references for all norms in $\mathcal{T}$.
For each norm, we prompt its content to GPT-4 and ask about the relationship between references inside the node attribute and the specified norm.
Each reference is then categorized as either a support or an exception of the given norm.

\subsection{Auxiliary Knowledge Collection}
In addition to norm annotation, most regulations also bring their own terminologies and use these terms throughout their content.
To facilitate the reasoning process and align the given context with corresponding regulations, we further collect two auxiliary graphs and one definition dictionary.

\paragraph{Role graph $\mathcal{G}^r$.}
Role instances extracted from daily context frequently misalign with the abstract roles defined in regulations.
For example, in HIPAA, the term covered entity may refer to 1) a health plan, 2) a health care clearinghouse or 3) a health care provider.
To permit Alice's action in Figure~\ref{fig:intro}, we need to link Alice's role, surgeon, to the covered entity.
Based on this observation, we further curate a role graph $\mathcal{G}^r = \{\mathcal{V}^r, \mathcal{E}^r, \mathcal{R}^r\}$ where vertices $v \in \mathcal{V}^r$ represent social roles and only two relations ``subsume'' and ``is subsumed by'' are in $\mathcal{R}^r$.
To construct our role graph $\mathcal{G}^r$, we align the roles in the legal documents with the WordNet~\cite{miller-1994-wordnet}.
We first use GPT-4 and parsers to recognize social roles in the legal document, then we align these roles in the WordNet and recursively append their hypernyms until the role root ``person.n.01'' is reached.
For new definitions of roles such as the covered entity, if these definitions include roles in $\mathcal{V}^r$, we further append these new definitions as roles.

\paragraph{Attribute graph $\mathcal{G}^a$.}
Similarly, to identify whether the transmitted attributes are sensitive, we need
an attribute graph $\mathcal{G}^a = \{\mathcal{V}^a, \mathcal{E}^a, \mathcal{R}^a\}$
where vertices $v \in \mathcal{V}^a$ represent private attributes and only two relations ``subsume'' and ``is subsumed by'' are in $\mathcal{R}^r$.
With $\mathcal{G}^a$, we can map the surgery operative report to the protected health information (PHI) specified by HIPAA. 
To collect private attributes protected by existing regulations, we incorporate expert annotated ontologies, including Data Privacy Vocabulary~\cite{Pandit-2019-DPV} and OPPO~\cite{gupta2023oppo} into $\mathcal{G}^a$.
Our $\mathcal{G}^a$ gathers the ontologies' inclusion relationships for both class-class and class-individual.
Then, for new terms that include attributes in $\mathcal{V}^a$, we further append them into $\mathcal{G}^a$.

\paragraph{Definition dictionary $\mathcal{D}$.}
To further facilitate in-context reasoning, we also implement an HTML parser to build a definition dictionary $\mathcal{D}$ based on legal documents' definition chapters.
Keys of $\mathcal{D}$ are introduced terms and terms' corresponding definitions are values.

\subsection{Data Summary and Manual Validation}

\paragraph{Data statistics.}
In our collected \name, our processed document tree $\mathcal{T}$ includes 642 internal nodes, 1,673 leaf nodes, 2,098 edges for subsumption relationship and 761 cross references.
For the annotated privacy rules, we identify 231 positive norms (permitted information transmission norms), 32 negative norms as well as 328 general definitions of specifications for applying norms within $\mathcal{T}$.
For our role graph $\mathcal{G}^r$, we collect 296 roles with 318 subsumption relations.
For our attribute graph $\mathcal{G}^a$, we collect 211 sensitive attributes with 485 subsumption relations.
Moreover, our definition dictionary $\mathcal{D}$ encompasses 54 definitions formally defined in the HIPAA 164.103.

\paragraph{Expert verification.}
To ensure the quality and correctness of our \name, two authors and one law student manually validate positive and negative norms with the original regulations and rectify the inconsistently parsed CI characteristics.

\section{Enhance Reasoning with Checklist}

In this section, we show how to exploit LLMs' in-context reasoning ability to determine privacy violations.
As shown in Figure~\ref{fig:model} (b), the basic intuition is to use existing LLMs for a retrieval-augmented generation (RAG) pipeline.

\subsection{Context Annotation Methods}
In addition to directly feeding the whole context for retrieval, we adopt the following annotation methods for a given context:

\paragraph{CI characteristics extraction.}
We follow the same CI characteristics extraction procedures used for building our checklist to extract the sender role, receiver role, subject role, transmitted attribute, purpose, and consent form from the given context.

\paragraph{LLM explanation.}

There are often discrepancies between the CI characteristics extracted from legal documents and those derived from real-world contexts. 
Typically, CI characteristics from legal documents are defined using exclusive legal terminology, whereas those from daily contexts tend to be more specific and practical.
Without the knowledge about definitions in $\mathcal{D}$, existing retrieval methods struggle to accurately search for relevant rules based on the given context.
For example, both statistical and semantic methods of information retrieval fail to link the given role query ``surgeon'' to ``covered entity''.
To address this, we leverage LLMs to explain the context using terms inside designated laws.
After the explanation, the retrieval process can be more accurate.

\subsection{Rule Retrieval Methods}

To retrieve relevant regulations from our \name, we implement the following retrieval methods based on the parsed context:

\paragraph{BM25.}
BM25 retrieves documents based on the term frequency-inverse document frequency (TF-IDF) score \cite{salton1988term} and word frequency. We use the LLM explanation as the query \( Q \), aiming to select the most relevant regulations by measuring the similarity between the LLM explanation query \( Q \) and the regulation texts \( E \) in the document tree \( \mathcal{T} \).
For a word \( w \) in the query \( Q \), the relationship between the word and the regulation definition is calculated as follows:
\begin{equation}
\small
    s(w,E) = \frac{{\rm IDF}(w) \cdot f(w, E) \cdot (k_1 + 1)}{f(w, E) + k_1 \cdot (1 - b + b \cdot \frac{|E|}{{\rm avgdl}})} , 
\end{equation}
where \( {\rm IDF}(w) \) represents the inverse document frequency of the word \( w \), \( f(w, E) \) denotes the frequency of the word \( w \) appearing in \( E \), \( |E| \) is the length of \( E \), and \( {\rm avgdl} \) is the average length across the subrules in \( \mathcal{T} \). 
$k_1$ and $b$ are hyperparameters, and we set $k_1 = 1.5$ and $b = 0.75$. 
The similarity between a query \( Q \) and the definition \( E \) is calculated by summing the scores of words in \( Q \):
\begin{equation}
\small
    Sim(Q,E) = \sum\nolimits_{w_i \in Q} s(w_i, E).
\end{equation}
The regulations with the highest scores are then selected as the retrieved results.

\paragraph{Embedding similarity.}
In addition to statistical similarity, we also implement retrieval methods based on semantic similarity.
Specifically, we use pretrained Sentence Transformers~\cite{reimers-2019-sentence-bert} to calculate both contexts and regulations' embeddings. 
Then, the cosine similarities are calculated for embeddings of the contexts and regulations.
Notably, besides calculating semantic similarity using chunks of text, we can also calculate semantic similarity on roles and attributes.
These roles and attributes can then be used as keywords to retrieve applicable rules.
For simplicity, we use ``all-mpnet-base-v2''~\cite{song-2020-mpnet} as our embedding model throughout the paper.

\paragraph{Agent-based retrieval.}

Besides directly retrieving relevant rules from our \name, we consider LLMs themselves as knowledge bases~\cite{petroni-etal-2019-language} that have preliminary legal knowledge.
Based on this heuristic, we implement agent-based retrieval by treating powerful LLMs as lawyers
and use prompt engineering to generate applicable rules with corresponding IDs.
To mitigate misinformation and hallucination, the agent-based retrieval further leverages our \name to verify if the generated IDs are valid.

\subsection{In-context Reasoning}

With the retrieved rules, we may leverage existing LLMs to perform in-context reasoning.
For each candidate rule, we apply a Chain-of-Thought (CoT) prompting template~\cite{Wei2022ChainOT, wang2023selfconsistency} to assess its applicability in judging privacy violations within a given context. 
Our CoT prompt template includes four components.
First, the LLM is instructed to identify the information transmission flow and analyze the three stakeholders, as well as the transmitted attributes and purpose from the given context.
Second, the LLM reviews the analyzed information and determines if it matches the candidate rule's legal terms.
Third, the LLM assesses whether the candidate rule is relevant to the context.
Lastly, the LLM decides whether the context is in compliance, in violation, or unrelated to the candidate rule.

\section{Experimental Setups}
\label{sec:exps}


\paragraph{Data.}
We use collected real and synthetic court cases for evaluation to demonstrate the usefulness of our \name.
Following~\citet{fan2024goldcoin}, we use their collected real and synthetic court cases about HIPAA.
The real cases are collected from the
Caselaw Access Project (CAP)\footnote{\url{https://case.law/}} and the synthetic cases are LLM-augmented data based on real cases and HIPAA sub-rules.
For both datasets, we directly prompt existing LLMs to perform three-way classification by determining if the given case is \textit{permitted} by, \textit{prohibited} by or \textit{not applicable} to HIPAA.
The dataset statistics are summarized in 
Table~\ref{tab:dataset-table}.

\paragraph{Evaluated LLMs.}
We evaluate both open-source and close-source LLMs.
In terms of open-source LLMs, we download their model weights of instruction or chat versions and generate responses on two NVIDIA Tesla A100 40GB graphic cards.
Our evaluated open-source models include  Llama3~\cite{llama3modelcard}, Qwen1.5, Qwen2~\cite{Yang2024Qwen2TR}, ChatGLM4~\cite{glm2024chatglm}, and Mistral-v0.3~\cite{jiang2023mistral}.
For close-source LLM, we evaluate the GPT-4-turbo-04-09's performance with API accesses.

\begin{table*}
\centering
\small
\begin{tabular}{l c c c c c}
\toprule
Type & Permit & Prohibit & Not Applicable & Avg Context Length & Avg Reference \#\\
\midrule
Real & 87 & 20 & 107 & 311.87 & 4.64\\
Synthetic  & 269 & 40 & 309 &187.30 & 1.09 \\

\bottomrule
\end{tabular}
\vspace{-0.05in}
\caption{\label{tab:dataset-table}
Statistics of the evaluation datasets.
}
\vspace{-0.1in}
\end{table*}

\begin{table*}[t]
\centering
\small
  \begin{tabular}{l c | c c c c c c }
    \toprule
      & {Type} & {DP} & {CoT-auto} & {CoT-manual} & {Agent-ID} & {BM25-content} & {CI-ES-content}\\
      \midrule
    
    \multirow{2}{*}{Llama3-instruct-8b}
    & Real & 77.57 & 79.43 & 72.89 & 86.44 & \textbf{87.85} & 85.98  \\
    & Synthetic & 82.52 & 93.52 & 94.49 & 94.49 & \textbf{95.46} & 95.30 \\
    \midrule
    \multirow{2}{*}{Qwen1.5-14b}
    & Real & 35.98 & \textbf{87.38} & 78.50 & 81.77 & 85.04 & 83.17  \\
    & Synthetic & 48.86 & \textbf{96.27} & 95.46 & 94.26 & 95.46 & 94.98 \\
    \midrule
    \multirow{2}{*}{Qwen2-7b}
    & Real & 48.13 & 68.69 & 63.55 & 71.02 & 67.75 & \textbf{79.44}  \\
    & Synthetic & 64.23 & 81.55 & 79.77 & 80.90 & 82.52 & \textbf{88.67} \\
       \midrule
    \multirow{2}{*}{GLM-4-chat-9b}
    & Real & 64.95 & 70.09 & 73.83 & 77.10 & \textbf{82.71} & 76.63  \\
    & Synthetic & 89.48 & 94.17 & \textbf{95.30} & 91.90 & 91.74 & 94.01 \\
       \midrule
    \multirow{2}{*}{Mistral-v0.3-7b}
    & Real & 60.28 & 64.01 & 63.55 & \textbf{69.62} & 69.15 & \textbf{69.62}  \\
    & Synthetic & 85.59 & 82.68 & 92.07 & 92.07 & \textbf{92.23} & 90.27 \\
       \midrule
    \multirow{1}{*}{GPT-4-turbo-04-09}
    & Real & 86.91 & 74.76 & 88.31 & 89.25 & \textbf{89.71} & 86.91  \\
        \midrule
    \multirow{2}{*}{\textbf{Average}}
    & Real & 62.30 & 74.06 & 73.43 & 79.20 & \textbf{80.36} & 80.29  \\
    & Synthetic & 74.13 & 89.63 & 91.41 & 90.72 & 91.48 & \textbf{92.64} \\
    
    \bottomrule
  \end{tabular}
  \vspace{-0.1in}
  \caption{\label{tab:overall}
The overall accuracy evaluation results. The accuracies are reported in\%.
}
\vspace{-0.2in}
\end{table*}

\paragraph{Evaluated methods.}
We evaluate LLMs' performance over the following methods. 
The first three methods rely on prompting tricks while the last three methods reason over our \name with different retrieval methods.

$\bullet$ Direct prompt (\textbf{DP}).
We prompt LLMs with the case context and directly ask them to determine if the given context is permitted, prohibited, or unrelated to HIPAA.

$\bullet$ CoT prompt with automatic planning (\textbf{CoT-auto}).
We prompt LLMs to automatically generate step-by-step planning to analyze the given case and then execute the steps to determine privacy violations.

$\bullet$ CoT Prompt with manual guidelines (\textbf{CoT-manual}).
Instead of using LLMs to generate plans, we prompt LLMs with pre-defined guidelines for each step.
Our guidelines follow the CI theory to instruct LLMs to analyze the CI characteristics step by step to assess privacy violations. 

$\bullet$ CoT prompt with regulation IDs from \textit{agent-based retrieval} (\textbf{Agent-ID}).
We first ask LLMs with the case to generate applicable regulation IDs.
Then, we verify their existence with our checklist to filter out misinformation.
Finally, we prompt LLMs similarly to the \textit{CoT-manual} approach, but with the addition of regulation IDs for in-context reasoning.


$\bullet$ CoT prompt with \textit{LLM explanation} and regulations retrieved via BM25 (\textbf{BM25-content}).
We apply the \textit{LLM explanation} to clarify the case context with legal terms to facilitate the retrieval process.
Then, we use BM25 to search for relevant sub-rules.
Finally, we incorporate both content and IDs of these sub-rules into the \textit{CoT-manual} prompt to improve in-context reasoning.

$\bullet$ CoT prompt with \textit{CI characteristics extraction} and regulations retrieved via embedding similarity (\textbf{CI-ES-content}).
We first perform \textit{CI characteristics extraction} to extract necessary information about information flow.
Then, we consider stakeholders' roles as keywords and use pre-trained embedding models to match their roles with roles in our \name by calculating the embedding similarities.
We use the matched roles as keywords to search for applicable rules.
Finally, we add the regulations to the \textit{CoT-manual prompt} for in-context reasoning.

\paragraph{Evaluation metrics.}
For the three-way classification, we report the accuracy of correct LLM judgments in a single run.
In addition, for each class, we also calculate its precision, recall, and F1.
To verify the correctness, we implement multiple parsers based on regular expressions to capture the desired predictions. 
All parsing failures are regarded as incorrect predictions.

\section{Experimental Results}
In this section, we show how our \name enhances existing LLMs' privacy assessment ability via zero-shot in context reasoning.

\subsection{Overall Assessments}
In this section, we evaluate the overall accuracy of the proposed six types of prompts over various LLMs.
Table~\ref{tab:overall} reports the accuracies for both \textit{real} and \textit{synthetic} court cases and includes six prompts' average performance.
According to the results, we summarize the following findings:

1) \textit{Our proposed \name is effective in improving LLMs' in-context reasoning ability to decide privacy violations.}
In general, \textit{agent-ID},  \textit{BM25-content} and \textit{CI-ES-content} outperform \textit{DP},  \textit{CoT-auto} and \textit{CoT-manual}, leading to around 6\% - 18\% accuracy improvements.
These improvements can be attributed to the retrieved sub-rules from our \name.

2) In terms of prompting tricks, we observe that \textit{manually crafted prompts based on contextual integrity theory leads to improved LLM performance.}
In our experiments, \textit{CoT-manual} generally outperforms \textit{DP} and  \textit{CoT-auto}.
Despite \textit{CoT-auto} also exploits LLMs' planning abilities to decompose the query into step-by-step plans, its automatic planning prompts still underperform \textit{CoT-manual}'s expert taxonomies by approximately 5\%.
This finding suggests that contextual integrity theory is beneficial for LLMs to reason about privacy violations.

3) \textit{LLMs perform better at synthetic data rather than real court cases.}
Even though the synthetic data is augmented via GPT-4, all open-source LLMs lead to significant accuracy improvements over all six prompting methods.
We manually inspect the synthetic samples and observe that LLMs tend to simplify the context to overfit the given instructions.
The context information of most synthetic data shifts to naive and unrealistic narratives.
Consequently, the distribution of synthetic data is dominated by easy-to-solve questions. 
Such distribution shift is also observed on LLM-augmented math problems~\cite{tong2024dartmath}.

4) \textit{LLMs still need improvements to serve as responsible judges.}
According to Table~\ref{tab:overall}'s results, all LLMs struggle to reach 90\% accuracy for real court cases.
Under the same prompt template, the results are highly influenced by LLMs' inherent reasoning and context-understanding ability.
Powerful foundation models, such as GPT-4, yield the best performance for real court cases.




\begin{table*}[!htbp]
\centering
\small
  \begin{tabular}{l  | ccc | ccc | ccc}
    \toprule
    \multirow{2}{*}{} &
    \multicolumn{3}{c|}{Permit} &
      \multicolumn{3}{c|}{Prohibit} &
      \multicolumn{3}{c}{Not Applicable} 
 \\
      & {Precision} & {Recall} & {F1} &{Precision} & {Recall} & {F1} &  {Precision} & {Recall} & {F1}\\
      \midrule
    
    \multirow{1}{*}{DP}
    & 87.67 & 73.56 & 80.00 & 45.83 & 55.00 & 50.00 & \textbf{97.84} & 85.04 & 91.00  \\
    \multirow{1}{*}{CoT-auto}
    & 86.36 & 65.51 & 74.50 & 27.27 & 60.00 & 37.50 & 97.11 & 94.39 & 95.73 \\
    \multirow{1}{*}{CoT-manual}
    & 84.09 & 42.52 & 56.48 & 22.41 & \textbf{65.00} & 33.33 & 95.49 & 99.06 & 97.24 \\
    \multirow{1}{*}{Agent-ID}
    & 89.47 & 78.16 & 83.43 & 52.63 & 50.00 & 51.28 & 90.67 & \textbf{100.00} & 95.11  \\
    \multirow{1}{*}{BM25-content}
    & 87.05 & \textbf{85.05} & \textbf{86.04} & \textbf{60.00} & 45.00 & \textbf{51.42} & 92.92 & 98.13 & \textbf{95.45}  \\
    \multirow{1}{*}{CI-ES-content}
    & \textbf{91.66} & 75.86 & 83.01 & 45.83 & 55.00 & 50.00 & 90.67 & 100.00 & 95.11  \\
        \midrule
    \multirow{1}{*}{Average}
    & 87.72 & 70.11 & 77.24 & 42.33 & 55.00 & 45.59 & 94.12 & 96.10 & 94.94 \\
    \bottomrule
  \end{tabular}
  \vspace{-0.1in}
  \caption{\label{tab:precision-recall}
The detailed investigation of Llama-3's performance over 3 classes on the real court cases.
Results are reported in \%.
}
\vspace{-0.15in}
\end{table*}


\begin{figure*}[t]
\centering
\includegraphics[width=0.99\textwidth]{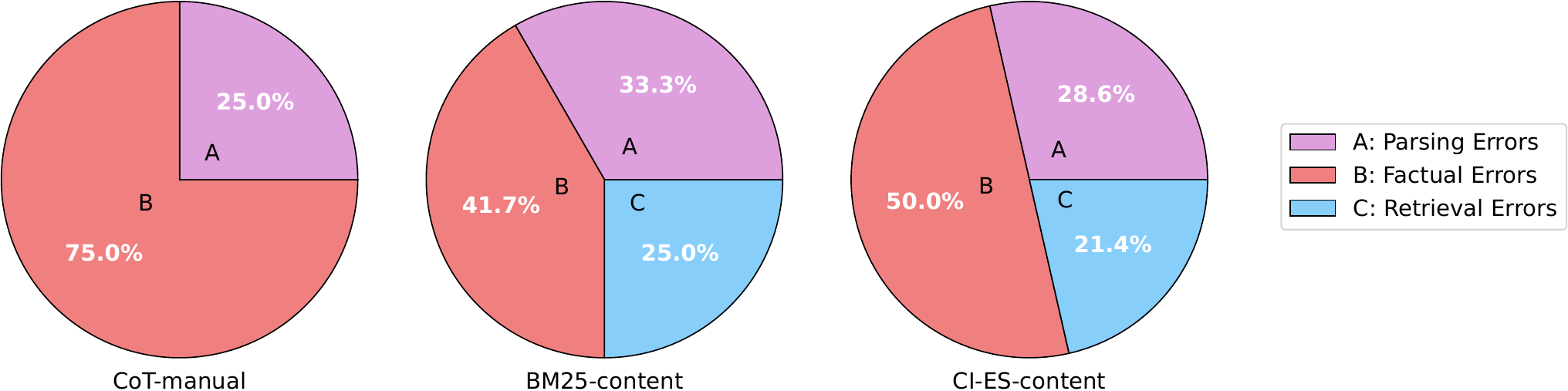}
\caption{
Manual investigations on GPT-4's errors for prohibited cases.
}
\label{fig:human_eval}
\vspace{-0.2in}
\end{figure*}

\subsection{Inspections on Class-level Performance}

In addition to evaluations of the overall accuracies, we also study the precision, recall and F1 within each class.
Without loss of generality, we examine the detailed performance of one representative LLM, Llama-3-instruct-8b, for real court cases over the six prompts.
Its precision, recall and F1 for each class are reported in Table~\ref{tab:precision-recall}.
Our results imply the following observations:

5) \textit{LLMs are capable of determining applicability and perform badly at prohibited cases.}
Cases that are not applicable achieve over 90\% F1 score for all prompts.
In contrast, permitted cases have an average F1 of 77.24\% and prohibited cases only have an average F1 of 45.59\%.
These results suggest that all LLMs are impotent and biased judges on prohibited cases even if their contexts are given.

6) \textit{CoT prompting only improves LLMs' performance on applicability, while our checklist helps LLMs to make correct judgments on permitted cases.}
Though both CoT prompts lead to improved overall performance, such improvements mainly come from increased F1 on applicability.
On the other hand, their F1 scores on both permitted and prohibited cases drop significantly.
CoT prompting fails to enhance LLMs to determine compliance for applicable cases.
Instead, the RAG-based pipelines not only achieve comparable performance on applicability as CoT prompting, but also improve LLMs' performance on permitted cases.

\subsection{Error Analysis via Human Evaluations}
In this section, we conduct a qualitative analysis to manually inspect LLMs' erroneous predictions.
Two authors and one law student examine 60 prohibited cases' logs of \textit{CoT-manual}, \textit{BM25-content} and \textit{CI-ES-content}  generated by GPT-4.
Upon investigation, we summarize three types of errors.
The first is the parsing error, where our parsers fail to capture the correct predictions.
The second is factual errors, which include hallucinations and misinterpretations of the context.
The last is retrieval errors whose retrieved rules are irrelevant and misleading.
Figure~\ref{fig:human_eval} depicts the error distributions for the three RAG-based prompts.
\lhr{
Since \textit{CoT-manual} is not based on the RAG pipeline, so its retrieval error is 0\%.
}

According to the pie charts, though GPT-4 is one powerful foundation model, its factual errors still constitute the most significant portion.
Such factual errors are commonly caused by hallucinations where LLMs reason on non-existing assumptions.
On the other hand, though prompting with our \name can reduce the rate of factual errors, more than 20\% of errors are caused by the irrelevant retrieved sub-rules.
These irrelevant sub-rules may further misguide LLMs in fitting their criteria and therefore generate far-fetched explanations.
This observation implies that LLMs' performance can be further improved by reducing hallucinations and enhancing the retriever's functionality.

\section{Conclusions and Future Works}

In this paper, we follow the spirit of Contextual Integrity Theory to transform privacy-related issues into reasoning problems.
Unlike prior attempts which manage to build rigorous expert systems based on logic languages, we exploit powerful LLMs to further formulate the reasoning problems to the in-context reasoning problems based on the retrieval-augmented generation (RAG) pipeline.
To facilitate the retrieval process, we propose \name, a checklist that covers processed legal documents, annotated CI characteristics,  and auxiliary knowledge about roles, private attributes and term definitions.
We compare three RAG-based prompts with three baseline prompts to show that our proposed \name can improve LLMs' privacy awareness for real and synthetic court cases.
Our future works can be divided into two folds.
First, we plan to integrate most mainstream privacy regulations as well as AI safety standards into our \name.
Second, we will work on improving the retrieval performance to eliminate the gap between common sense knowledge and legal terminologies.
\lhr{After getting over the two parts, we may no longer need the privacy-utility trade-off for model training and inference.
Instead, we may simply apply our checklist as a detector for safeguarding models' inputs and outputs and concentrate on models' utility improvements.}



\section*{Acknowledgements}

The authors of this paper were supported by the ITSP Platform Research Project (ITS/189/23FP) from ITC of Hong Kong SAR, China, and the AoE (AoE/E-601/24-N), the RIF (R6021-20) and the GRF (16211520 and 16205322) from RGC of Hong Kong SAR, China.

\section*{Limitations}

When we apply our \name to LLMs to evaluate compliance with a given court case, a major limitation is that LLMs cannot perform well on prohibited cases, even for GPT-4 models.
As mentioned in the experimental results of Table~\ref{tab:precision-recall}, LLMs are impotent and biased judges on prohibited cases.
After manually inspecting all prohibited cases, we find that LLMs are likely to assume that conditions not presented in the context, but required by the retrieved regulations, are satisfied.
Such hypotheses cause LLMs to hallucinate and further conclude that a positive norm is satisfied for prohibited cases.
This limitation suggests current LLMs are not ready yet to be legal agents.
Instead, we should further tune LLMs to align with given contexts without unnecessary ``if statements'' to use LLMs to detect privacy violations.

Additionally, since LLMs with 7-14 billion parameters usually have context lengths of no more than 8k,  we cannot feed few-shot demonstrations into these LLMs to evaluate the few-shot performance.

\section*{Ethical Considerations}
We declare that all authors of this paper acknowledge the \emph{ACM Code of Ethics} and honor the ACL code of conduct.
Our work claims that simply studying leakage of PII patterns may not be enough to align with people's actual concerns.
Instead, we consider the context-centric approach to determine privacy violations grounded on established privacy norms and standards.
We believe that our checklist will become a new paradigm for studying privacy issues as LLMs are consistently improving their reasoning abilities.

\textbf{Checklist Construction}. 
During the construction process of the \name, we parse the HIPAA document tree from the official website of the Code of Federal Regulations following their granted access rules.
In terms of expert verification steps, all manual annotations are initially done by two of the authors and another student with a law degree who can provide professional suggestions.

\textbf{Evluation Data}. 
Both synthetic and real court cases used in our evaluation are publicly available from GoldCoin's official GitHub implementation\footnote{\url{https://github.com/HKUST-KnowComp/GoldCoin}} under the Apache-2.0 license.
For the real court case collection procedure, GoldCoin follows the official usage and access rules of the Caselaw Access Project\footnote{\url{https://case.law/about/\#usage-access}} during downloading relevant cases.
\clearpage


\bibliography{custom}

\clearpage
\appendix

\section{Potential Impacts of Checklist}

In this section, we discuss the potential impacts of our proposed \name.

\subsection{Connections with Legal Reasoning}
American legal scholars commonly decompose legal reasoning into four sequential stages, including Issue, Rule, Application, and Conclusion (IRAC)~\cite{guha2023legalbench, wiki:irac}.
``Issue'' relates to issue-spotting to identify the legal issues in the given context.
``Rule'' aims to recall relevant legal rules for identified issues.
Then, lawyers apply these rules (``Application'') to analyze the rule applicability.
Lastly, lawyers reach their conclusions after the application stage.
Our checklist follows a similar spirit of IRAC to enable retrieval augmented generations for LLMs.
For the retrieval process, we actually complete the Issue and Rule stages.
Then, we leverage LLMs to finally handle the Application and Conclusion stages.
Unlike existing legal reasoning tasks that are confined to evaluating partial stages of IRAC~\cite{guha2023legalbench,pipitone2024legalbench}, our checklist enables LLMs to complete all 4 stages consecutively, which can be regarded as the ultimate objective of legal reasoning.

\subsection{Broad Use Cases}
Our checklist offers new solutions for the following 3 use cases.

1) A generative privacy breaches classifier.
Current safety alignment methods lead to over-defense problems and safety data covers approximately 20\% - 50\% alignment data.
Our checklist can be used to train an external guard to identify privacy-intruding model prompts and outputs to avoid potential harm.
Then, the internal model only needs to be aligned for utility improvements.

2) A general privacy consultant.
In addition, our checklist can serve as a privacy consultant for the public interest.
Users are able to receive in-time responses about their privacy concerns.

3) A detector for conflicting legal documents.
In the future, after we cover more regulations, our checklist can even be used to enumerate all possible roles and attributes to help legal scholars identify unnoticed conflicts between state laws and federal laws to determine preemption.

\section{Experimental Details}

\paragraph{Computational resources.} During our experiment, we use 2 NVIDIA RTX 6000 to run our codes, and it takes GPU hours around 2 weeks to complete all experiments.

\paragraph{Generation details.}
For open-source models, we generate the models' responses with the recommended configurations in their model cards.
For close-source models, we use their official APIs to obtain the responses with temperature = 0 and top\_p = 0.95 to ensure reproducibility.

\paragraph{Prompt templates for evaluation.}
In this paragraph, we list all prompt templates used to evaluate LLMs' performance.
The full prompts for the first three non-RAG methods including \textbf{DP}, \textbf{CoT-auto} and \textbf{CoT-manual} are shown in Table~\ref{tabs:non_rag_prompt}.
The remaining three RAG-based methods, generally follow the retrieval, filter, and decision-making pipeline.
For the retrieval stage, different retrieval methods are used to get relevant sub-rules.
For the filter stage, the same prompt can be applied to all methods to verify the retrieved sub-rule is relevant to the given event.
Finally, the decision-making stage follows previous chain-of-thought reasoning prompts by incorporating the selected reference regulations after the filter stage.
Prompts used for \textbf{Agent-ID} are presented in Table~\ref{tabs:prompt_agentid}. Those for \textbf{BM25-content} can be found in Table~\ref{tabs:prompt_BM25}, while Table~\ref{tabs:prompt_emb} displays the prompts used for \textbf{CI-ES-content}.

\paragraph{Licenses.}
For our evaluation data, we use data provided by GoldCoin's official GitHub implementation under the Apache-2.0 license.
In terms of used models, we follow all the licenses to use their models for the research propose.
For example, we follow Llama 3 Community License Agreement to use the Llama3-8b model for experiments.

\section{Privacy Leakage on ML Models}
\label{app: relate}

Recent studies on privacy issues can be categorized as training data leakage and inference data leakage.
For training data leakage, private data may be unintentionally collected for training and extracted via training data extraction attacks~\cite{carlini-2021-extracting, li2023multi,deng2023jailbreaker}.
In terms of inference data leakage, users' private data used during inference may also be recovered.
For example, sensitive system-level prompts may be extracted via prompt injection attacks~\cite{Wan2023Poisoning,ignore_previous_prompt,Liu2023PromptIA,shu2023exploitability} and side-channel attacks~\cite{debenedetti2023privacy}.
Moreover, information leakage~\cite{Gu2023TowardsSL, li-etal-2022-dont, Song-2020-Information} may also occur on models' hidden representation to recover sensitive information.

\section{CI Characteristics Extraction Details}
In this section, we give the detailed prompts used to extract CI characteristics for each leaf node in our parsed document tree $\mathcal{T}$.
Our full prompt is shown in Table~\ref{tabs: CI-extraction}.
We convert the CI characteristics extraction steps into a series of questions.
Then, we use regular regular expressions to parse answers for each question.
We re-run the prompt whenever we encounter a pattern-matching failure.

\section{Comparison with GoldCoin}

In this section, we compare our checklist with GoldCoin~\cite{fan2024goldcoin} in the following aspects to show our contributions.

\subsection{Usability}
Current best-performing LLMs are either close-sourced with only commercial API accesses or open-sourced with giant model scales (i.e., 405B parameters) that are hard to fine-tune.
Consequently, it is costly to apply GoldCoin to a wide range of models.
Instead, our \name offers plug-in solutions to enhance LLMs on privacy compliance tasks with better usability.

\subsection{Downstream Performance}
Since GoldCoin and our checklist evaluate the performance of different LLMs with different setups (GoldCoin decomposes the three-way classification into two binary classification tasks), we cannot make a fair comparison.

We follow the setting of the binary compliance task in GoldCoin to compare our RAG-based methods with GoldCoin in Table~\ref{tabs:goldcoin} on open-sourced LLMs. 
According to the overall Ma-F1 score, our BM25-content generally outperforms GoldCoin with the advanced LLMs.

In terms of close-sourced LLMs such as GPT-4, GoldCoin cannot fine-tune these LLMs to improve model performance.
Instead, our checklist enables RAG-based various methods to further improve close-sourced LLMs' performance, as shown in Table~\ref{tab:overall}.

\begin{table*}[!t]

    \renewcommand\arraystretch{1.06}
    \small
    \begin{center}       
    \begin{tabular}{m{2cm}|m{2cm}|m{0.95cm}<{\centering}m{0.95cm}<{\centering}m{0.95cm}<{\centering}|m{0.95cm}<{\centering}m{0.95cm}<{\centering}m{0.95cm}<{\centering}|m{0.95cm}<{\centering}m{0.95cm}}
            \toprule
            \multirow{2}{*}{\textbf{Method}} & \multirow{2}{*}{\textbf{Models}}& \multicolumn{3}{c|}{\textbf{Permit}} &  \multicolumn{3}{c|}{\textbf{Forbid}}  & \multicolumn{1}{c}{\textbf{All}} \\
            && \textbf{Prec} & \textbf{Rec} & \textbf{F1}& \textbf{Prec} &  \textbf{Rec} & \textbf{F1} &  \textbf{Ma-F1}  \\
            \midrule
            \multirow{4}{*}{{\textbf{GoldCoin}}}
            &MPT-7B     &86.49&73.56&79.50&30.30&50.00&37.74&58.62\\ 
            &Llama2-7B   &84.21&91.95&\textbf{87.91}&41.67&25.00&31.25&59.58\\
            &Mistral-7B   &90.67&78.16&83.95&40.62&65.00&50.00&66.98\\
            &Llama2-13B    &87.80&82.76&85.21&40.00&50.00&44.44&64.83\\

            \midrule
            \midrule

            \multirow{1}{*}{\textbf{Agent-ID}} & Llama3-8B  
    & 89.47 & 78.16 & 83.43 & 52.63 & 50.00 & 51.28 & 67.36\\
    \multirow{1}{*}{\textbf{BM25-content}} & Llama3-8B  
    & 87.05 & 85.05 & 86.04 & 60.00 & 45.00 & \textbf{51.42}  & \textbf{68.73}\\
    \multirow{1}{*}{\textbf{CI-ES-content}} & Llama3-8B  
    & 91.66 & 75.86 & 83.01 & 45.83 & 55.00 & 50.00 &  66.50\\
    
    \bottomrule
    \end{tabular}
    \end{center}
\caption{Performance comparison between GoldCoin and our RAG-based methods. Ma-F1 denotes the macro-F1 score.}
\label{tabs:goldcoin}
\end{table*}

\begin{table*}[h]
\centering
\small

\begin{tabular}{p{0.96\textwidth}}
\toprule
As a legal expert specializing in the HIPAA Privacy Rule, your task is to read a specific paragraph of the regulation \\
\textcolor{contentcolor}{<Regulation ID>}:
\newline
\textcolor{contentcolor}{<Regulation Content>}

Now complete the following questions one by one:
\newline
\newline
Q1. ("Prohibit", "Permit" or "General Definition") Ascertain whether the regulation \textcolor{contentcolor}{<Regulation ID>} pertains to scenes that are:\\
A. Prohibit by law\\
B. Permit by law\\
C. General Definition
\newline
\newline
Q2. (Identify the stakeholders related to regulation \textcolor{contentcolor}{<Regulation ID>}) Identify the stakeholders related to the regulation \textcolor{contentcolor}{<Regulation ID>}. Your response must include the following seven characteristics about the flow of private information: [Sender, Sender Role, Recipient, Recipient Role, Subject, Subject Role, Information Type, Consent Form, Purpose]. Answer 'None' if no information about characteristics is present.

The "Sender," "Recipient," and "Subject" ﬁelds indicate the sender, recipient, and the data subject during information transmission.\\
The "Sender Role", "Recipient Role" and "Subject Role" ﬁelds indicate the role of the sender, recipient and subject (e.g., doctor, patient).\\
The "Information Type" ﬁeld deﬁnes what kind of information would be passed, such as name or location.\\
The "Consent Form" ﬁeld indicates whether the sender has obtained consent from the subject to send the message. If consent is required, you should answer "Consent" for a flexible requirement or "Authorization" for a formal and mandatory process required by the context. If consent is not related to the context, you should answer "None."\\
The "Purpose" ﬁeld indicates the purpose of the mentioned information transmission, such as treatment, payment, or health care operations. 
\newline
\newline
Q3: Are the Sender and Subject the same person?\\
A. Yes\\
B. No\\
C. Not Sure
\newline
\newline
Q4: Is Recipient and Subject the same person?\\
A. Yes\\
B. No\\
C. Not Sure
\newline
\newline
Q5. (Identify the relation between \textcolor{contentcolor}{<Regulation ID>} other sub-parts referred to in the context) Identify the relation between \textcolor{contentcolor}{<Regulation ID>} and referred \textcolor{contentcolor}{<Regulation Reference List>}. For each reference in \textcolor{contentcolor}{<Regulation Reference List>}, according to the context given, determine if the reference is an exception of \textcolor{contentcolor}{<Regulation ID>} or support the argument of \textcolor{contentcolor}{<Regulation ID>}. Answer with "Exception" or "Support" for each reference.\\
\bottomrule
	\end{tabular}
\caption{Our CI characteristics extraction prompt. Light blue texts inside each ``\textcolor{contentcolor}{<>}'' block denote a string variable.}
\label{tabs: CI-extraction}
\end{table*}

\section{Comparison between Synthetic and Real Data}

In this section, we provide quantitative and manual analyses to illustrate why our checklist performs worse on the real data.

For the quantitative study, as shown in Table~\ref{tab:dataset-table}, we compare the average word numbers that a case contains (Avg Context Length) and average reference numbers that a given case is relevant to (Avg Reference \#). 
The reference refers to the applicable norms. According to the result, it is obvious that real court cases have longer context lengths and reference numbers than synthetic cases, which suggests that real court cases are more complex.

For the manual evaluation, we randomly selected 50 samples from both synthetic and real data, including permitted and prohibited cases. 
For each pair of (real, synthetic) samples, we then analyzed and compared their contexts based on three key factors: complexity, authenticity, and consistency.
Complexity assesses which vignette appears more complicated or detailed.
Authenticity evaluates which story feels more realistic or genuine.
Consistency determines which story maintains better internal coherence.
We invite two postgraduate students with law degrees to consider which context is better regarding the three factors for the 50 sampled pairs and report the averaged results. As shown in Table~\ref{tab:app-compare}, the results suggest that real court cases are much more complex, authentic, and consistent than synthetic cases. Even though synthetic cases look plausible, real cases have much more detailed and complex backgrounds. Moreover, real cases do not use the exact terminologies used for the HIPAA. Hence, real cases are more difficult to handle for LLMs.

\begin{table*}[t]
\centering
\small
  \begin{tabular}{l  | ccc | ccc | ccc}
    \toprule
    \multirow{2}{*}{} &
    \multicolumn{3}{c|}{Complexity} &
      \multicolumn{3}{c|}{Authenticity} &
      \multicolumn{3}{c}{Consistency} 
 \\
      & {Win} & {Tie} & {Lose} &{Win} & {Tie} & {Lose} &  {Win} & {Tie} & {Lose}\\
      \midrule
    
    \multirow{1}{*}{Real}
    & 93.00 & 3.00 & 4.00 & 85.00 & 2.00 & 13.00 & 57.00 & 40.00 & 3.00  \\
    \multirow{1}{*}{Synthetic}
    & 4.00 & 3.00 & 93.00 & 13.00 & 2.00 & 85.00 & 3.00 & 40.00 & 57.00 \\
    
    \bottomrule
  \end{tabular}
  \vspace{-0.1in}
  \caption{\label{tab:app-compare}
Manual comparison between real and synthetic data.
Results are reported in \%.
}
\vspace{-0.15in}
\end{table*}

\begin{table*}[t!]

\small
\centering
\begin{tabular}{p{2\columnwidth}}
\toprule

\textbf{DP}\\
    Determine the relationship between the provided event and HIPAA regulations. You have three options for your conclusion:\\
(A) The event is prohibited by HIPAA.\\
(B) The event is permitted by HIPAA.\\
(C) The event is not related to HIPAA.\\

Event:
\textcolor{contentcolor}{<event>}

Output Format:

Choice: [A. Prohibited | B. Permitted | C. Not related ]
\\

\midrule
\textbf{CoT-auto} \\
Determine the relationship between the provided event and HIPAA regulations. You have three options for your conclusion:\\
(A) The event is prohibited by HIPAA.\\
(B) The event is permitted by HIPAA.\\
(C) The event is not related to HIPAA.\\

Event:
\textcolor{contentcolor}{<event>}

Let's approach this step by step:

Planning: Outline the steps (no more than 6) necessary to analyze the relationship of the event with HIPAA standards.\\
Execution: Implement these steps to gather information and assess the specific aspects of the event.\\
Decision Making: Based on the information collected and analyzed, determine if there was a HIPAA violation.\\
Output Format:

Plans: [List the steps planned to evaluate the event:]\\
plan 1 - ...\\
plan 2 - ...\\
....\\
plan N - ...\\

Execution: [Document the outcomes from executing your plans:]\\
plan 1 - ....\\
plan 2 - ....\\
....\\
plan N - ...\\

Choice: [A. Prohibited | B. Permitted | C. Not related ]\\
\midrule
\textbf{CoT-manual}\\
Determine the relationship between the provided event and HIPAA regulations. You have three options for your conclusion:\\
(A) The event is prohibited by HIPAA.\\
(B) The event is permitted by HIPAA.\\
(C) The event is not related to HIPAA.\\

Event:
\textcolor{contentcolor}{<event>}

Let's approach this step by step:

1. Identify Relevant HIPAA Regulations: Start by researching and compiling the HIPAA regulations that could be relevant to the event, particularly focusing on privacy, security, and breach notification rules.\\
2. Examine the Nature of Information: Determine if the event involves any Protected Health Information (PHI), such as names, medical records, or other personal identifiers.\\
3. Evaluate HIPAA Applicability: Assess whether the event actually falls under the jurisdiction of HIPAA based on the nature of the entity involved and the type of information handled.\\
4. Assess Authorization and Consent: If HIPAA is applicable, verify if the recipient of the information was authorized and check for valid patient consent or another legal basis for sharing PHI.\\
5. Evaluate Security Measures: If applicable, review the security measures used during the information transmission to ensure they comply with HIPAA's security protocols.\\

Output Format:

Execution: [Document the outcomes from executing your plans:]\\
plan 1 - ....\\
plan 2 - ....\\
....\\
plan 5 - ....\\

Choice: [A. Prohibited | B. Permitted | C. Not related ]\\

\bottomrule
\end{tabular}
\caption{Prompts used without retrieval augmented generation. Light blue texts inside each ``\textcolor{contentcolor}{<>}'' block denote a string variable.}
\label{tabs:non_rag_prompt}
\end{table*}

\begin{table*}[t!]

\small
\centering
\begin{tabular}{p{2\columnwidth}}
\toprule

\textbf{LLMs as Knowledge Bases to Retrieve Regulation IDs}.\\
Read the event described below and generate the applicable HIPAA regulations (no more than \{generated\_num\}). This regulation will assist in determining if the event violates HIPAA security principles in a downstream task.

Event: \textcolor{contentcolor}{<event>}

Let's complete it step by step:\\
1. Review the Event Details: Understand the specifics of the event, including the type of information sent, the method of transmission, and the parties involved.\\
2. Identify Key HIPAA Concerns: Based on the event, identify potential concerns related to privacy, security, and breach notifications.\\
3. Retrieve Relevant Regulations: Consult the HIPAA regulatory text to find sections specifically addressing the identified concerns. Consider feedback to avoid repeating previously rejected regulations.\\

Output Format:

Execution: [Document the outcomes from executing the steps:]\\
1. - ...\\
2. - ...\\
...\\
4. - ...\\

Generated Related HIPAA Regulations(e.g. re"[0-9]+\.[0-9]+(\textbackslash([0 - 9A-Za-zivx]+\textbackslash))*" - xxxx):\\
1. Section Number - Section Content\\
...\\
N. Section Number - Section Content\\

Output:\\
\midrule
\textbf{Law Filter Template}.\\
Given a regulation and an information sending or requesting event, identify if the given regulation is relevant to the event. The process involves the following steps:

1. **Understand the Event**:
   - Extract key details from the description of the event, such as the type of information being exchanged, the parties involved, and the context or domain (e.g., healthcare, finance, education).

2. **Analyze The Regulation**:
   - For given regulation, determine its scope and main focus by reading the summary or key sections. Identify the primary subject matter, applicable contexts, or targeted stakeholders.

3. **Relevance Matching**:
   - Compare the key details of the event with the main focus of the regulation. Consider:
     - Does the regulation explicitly address scenarios similar to the event?
     - Is the regulation intended for the context or domain of the event?
     - Are the parties involved in the event the type of entities the regulation aims to govern?

4. **Filtering Decision**:
   - If a regulation's focus aligns well with the event’s details, please answer yes.
   - If there is little to no alignment, such as different contexts, unrelated subjects, or inappropriate scopes, please answer no.

Event: \textcolor{contentcolor}{<event>}

HIPAA Regulation:
\textcolor{contentcolor}{<candidates>}

Note:
It is possible that the event may be completely unrelated to the HIPAA regulations provided. In such cases, just answer NONE.

Question:
Is the given HIPAA Regulation relevant with the given event?

Output Format:
First answer yes or no, then explain the reason based on the given steps.\\
\midrule
\textbf{Decision-making Template}.\\
Determine the relationship between the provided event and HIPAA regulations. You have three options for your conclusion:\\
(A) The event is prohibited by HIPAA.\\
(B) The event is permitted by HIPAA.\\
(C) The event is not related to HIPAA.

I will offer you some retrieved HIPAA regulations for reference(Not 100\% correct.) Consider the specifics of the event, offered sections of HIPAA regulations.

Event:
\textcolor{contentcolor}{<event>}

HIPAA Regulations Reference:
\textcolor{contentcolor}{<reference\_regulations>}

Let's complete it step by step:\\
1. Understand the Event: Read the description of the event carefully to know exactly what happened.\\
2. Look Up HIPAA Rules: Get the HIPAA regulations that are provided and find the parts that might relate to the event.\\
3. Check for Key Points: Focus on important details of the event like the kind of information involved, who is handling it, and how it's being shared or used.\\
4. Compare the Event with the Rules: See how the details of the event stack up against the HIPAA rules to find any matches or issues.

Output Format:

Execution: [Document the outcomes from executing the steps:]\\
1. - ...\\
2. - ...\\
...\\
4. - ...\\

Choice: [A. Prohibited | B. Permitted | C. Not related ]\\
\bottomrule
\end{tabular}
\caption{Prompts used for \textbf{Agent-ID}. Light blue texts inside each ``\textcolor{contentcolor}{<>}'' block denote a string variable.}
\label{tabs:prompt_agentid}
\end{table*}

\begin{table*}[t!]

\small
\centering
\begin{tabular}{p{2\columnwidth}}
\toprule
\textbf{LLM Explanation for BM25 Similarity}.\\
I will provide you with an event concerning the delivery of information. Your task is to generate content related to this event by applying your knowledge of HIPAA regulations.

To ensure the content is relevant and accurate, follow these steps:

1. Understand the Event: Clearly define and understand the specifics of the event. Identify the key players involved, the type of information being handled, and the context in which it is being delivered.\\ 
2. Apply HIPAA Knowledge: Utilize your understanding of HIPAA regulations, focusing on privacy, security, and the minimum necessary information principles. Ensure that your content addresses these aspects in the context of the event.\\ 

Event Details:
\textcolor{contentcolor}{<event>}

Output Format:

Execution:

1. Identify the key players, type of information, and context.\\ 
2. Apply relevant HIPAA principles to the event.

Generated HIPAA Content:\\ 
1. HIPAA Privacy Rule: ...\\ 
2. HIPAA Security Rule: ...\\ 
3. Minimum Necessary Standard:...\\ 

References:\\ 
List the specific HIPAA regulations you consulted to generate the content.\\
\midrule

\textbf{Law Filter Template}.\\
Given a regulation and an information sending or requesting event, identify if the given regulation is relevant to the event. The process involves the following steps:

1. **Understand the Event**:
   - Extract key details from the description of the event, such as the type of information being exchanged, the parties involved, and the context or domain (e.g., healthcare, finance, education).

2. **Analyze The Regulation**:
   - For given regulation, determine its scope and main focus by reading the summary or key sections. Identify the primary subject matter, applicable contexts, or targeted stakeholders.

3. **Relevance Matching**:
   - Compare the key details of the event with the main focus of the regulation. Consider:
     - Does the regulation explicitly address scenarios similar to the event?
     - Is the regulation intended for the context or domain of the event?
     - Are the parties involved in the event the type of entities the regulation aims to govern?

4. **Filtering Decision**:
   - If a regulation's focus aligns well with the event’s details, please answer yes.
   - If there is little to no alignment, such as different contexts, unrelated subjects, or inappropriate scopes, please answer no.

Event:\textcolor{contentcolor}{<event>}

HIPAA Regulation:
\textcolor{contentcolor}{<candidates>}

Note:
It is possible that the event may be completely unrelated to the HIPAA regulations provided. In such cases, just answer NONE.

Question:
Is the given HIPAA Regulation relevant with the given event?

Output Format:
First answer yes or no, then explain the reason based on the given steps.\\

\midrule
\textbf{Decision-making Template}.\\
Assess the relationship between the provided event and HIPAA regulations, and select one of the following conclusions:

(A) The event is prohibited by HIPAA.\\
(B) The event is permitted by HIPAA.\\
(C) The event is not related to HIPAA.\\

You will be given some sections of HIPAA regulations for reference (Note: these references may not be 100\% accurate). Your task is to consider the specifics of the event in light of these sections. Remember that the provided references are not exhaustive, and the presence of a regulation does not automatically imply relevance or violation.

Event:
\textcolor{contentcolor}{<event>}

HIPAA Regulations Reference:
\textcolor{contentcolor}{<reference\_regulations>}

Let's complete it step by step:\\
1. Analyze the specific details of the event by identifying who is involved (sender and recipient), what information is being sent or requested, and for what purpose.\\
2. Compare key elements of the event with HIPAA rules, identifying if they involve the use, disclosure, or protection of Protected Health Information (PHI) as defined by HIPAA.\\
3. Evaluate the provided HIPAA regulation excerpts to see if they directly relate to the event.\\
4. Consider if there are other HIPAA rules not mentioned in the excerpts that might apply.\\
5. Conclude based on the comprehensive analysis whether the event is in compliance, in violation, or unrelated to HIPAA.\\

Output Format:\\

Execution: [Document the outcomes from executing each step]:\\
1. - ...\\
2. - ...\\
...\\
5. - ...\\

Choice: [A. Prohibited | B. Permitted | C. Not related]\\
\bottomrule
\end{tabular}
\caption{Prompts used for \textbf{BM25-content}. Light blue texts inside each ``\textcolor{contentcolor}{<>}'' block denote a string variable.}
\label{tabs:prompt_BM25}
\end{table*}

\begin{table*}[t!]

\small
\centering
\begin{tabular}{p{2\columnwidth}}
\toprule

\textbf{Law Filter Template}.\\
Given a regulation and an information sending or requesting event, identify if the given regulation is relevant to the event. The process involves the following steps:

1. **Understand the Event**:
   - Extract key details from the description of the event, such as the type of information being exchanged, the parties involved, and the context or domain (e.g., healthcare, finance, education).

2. **Analyze The Regulation**:
   - For given regulation, determine its scope and main focus by reading the summary or key sections. Identify the primary subject matter, applicable contexts, or targeted stakeholders.

3. **Relevance Matching**:
   - Compare the key details of the event with the main focus of the regulation. Consider:
     - Does the regulation explicitly address scenarios similar to the event?
     - Is the regulation intended for the context or domain of the event?
     - Are the parties involved in the event the type of entities the regulation aims to govern?

4. **Filtering Decision**:
   - If a regulation's focus aligns well with the event’s details, please answer yes.
   - If there is little to no alignment, such as different contexts, unrelated subjects, or inappropriate scopes, please answer no.

Event:\textcolor{contentcolor}{<event>}

HIPAA Regulation:
\textcolor{contentcolor}{<candidates>}

Note:
It is possible that the event may be completely unrelated to the HIPAA regulations provided. In such cases, just answer NONE.

Question:
Is the given HIPAA Regulation relevant with the given event?

Output Format:
First answer yes or no, then explain the reason based on the given steps.\\

\midrule
\textbf{Decision-making Template}.\\
Assess the relationship between the provided event and HIPAA regulations, and select one of the following conclusions:

(A) The event is prohibited by HIPAA.\\
(B) The event is permitted by HIPAA.\\
(C) The event is not related to HIPAA.\\

You will be given some sections of HIPAA regulations for reference (Note: these references may not be 100\% accurate). Your task is to consider the specifics of the event in light of these sections. Remember that the provided references are not exhaustive, and the presence of a regulation does not automatically imply relevance or violation.

Event:
\textcolor{contentcolor}{<event>}

HIPAA Regulations Reference:
\textcolor{contentcolor}{<reference\_regulations>}

Let's complete it step by step:\\
1. Analyze the specific details of the event by identifying who is involved (sender and recipient), what information is being sent or requested, and for what purpose.\\
2. Compare key elements of the event with HIPAA rules, identifying if they involve the use, disclosure, or protection of Protected Health Information (PHI) as defined by HIPAA.\\
3. Evaluate the provided HIPAA regulation excerpts to see if they directly relate to the event.\\
4. Consider if there are other HIPAA rules not mentioned in the excerpts that might apply.\\
5. Conclude based on the comprehensive analysis whether the event is in compliance, in violation, or unrelated to HIPAA.\\

Output Format:\\

Execution: [Document the outcomes from executing each step]:\\
1. - ...\\
2. - ...\\
...\\
5. - ...\\

Choice: [A. Prohibited | B. Permitted | C. Not related]\\
\bottomrule
\end{tabular}
\caption{Prompts used for \textbf{CI-ES-content}. Light blue texts inside each ``\textcolor{contentcolor}{<>}'' block denote a string variable.}
\label{tabs:prompt_emb}
\end{table*}

\end{document}